\documentclass[11pt,a4paper]{article}
\usepackage[hyperref]{emnlp2020}
\usepackage{times}
\usepackage{latexsym}
\usepackage{graphicx}
\usepackage{subcaption}
\usepackage{enumitem}
\usepackage{booktabs}
\usepackage{bm}
\usepackage{amsmath}
\usepackage{txfonts}

\def\1{\bm{1}}
\def\vx{{\bm{x}}}
\def\vy{{\bm{y}}}
\def\vz{{\bm{z}}}
\newcommand{\ie}{\textit{i}.\textit{e}., }
\newcommand{\eg}{\textit{e}.\textit{g}., }

\definecolor{mygray}{gray}{0.6}

\usepackage{microtype}
\aclfinalcopy

\title{Generating Synthetic Data for Task-Oriented Semantic Parsing with Hierarchical Representations}

\author{Ke Tran \\
  Amazon Translate \\
  Berlin, Germany \\
  \texttt{trnke@amazon.com} \\\And
  Ming Tan \\
  Amazon Alexa \\
  Cambridge, MA, USA \\
  \texttt{mingtan@amazon.com} \\}

\date{}

\begin{document}
\maketitle
\begin{abstract}
  Modern conversational AI systems support natural language understanding for a wide variety of capabilities. While a majority of these tasks can be accomplished using a simple and flat representation of intents and slots, more sophisticated capabilities require complex hierarchical representations supported by semantic parsing. State-of-the-art semantic parsers are trained using supervised learning with data labeled according to a hierarchical schema which might be costly to obtain or not readily available for a new domain. In this work, we explore the possibility of generating synthetic data for neural semantic parsing using a pretrained denoising sequence-to-sequence model (\ie BART). Specifically, we first extract masked \emph{templates} from the existing labeled utterances, and then fine-tune BART to generate synthetic utterances conditioning on the extracted templates. Finally, we use an auxiliary parser (AP) to filter the generated utterances. The AP guarantees the quality of the generated data. We show the potential of our approach when evaluating on the Facebook TOP dataset\footnote{\url{http://fb.me/semanticparsingdialog}} for navigation domain.
\end{abstract}

\section{Introduction}
\label{sec:intro}
\begin{figure}[ht]
  \centering
   \begin{minipage}[b]{0.47\textwidth}
    \includegraphics[width=\textwidth]{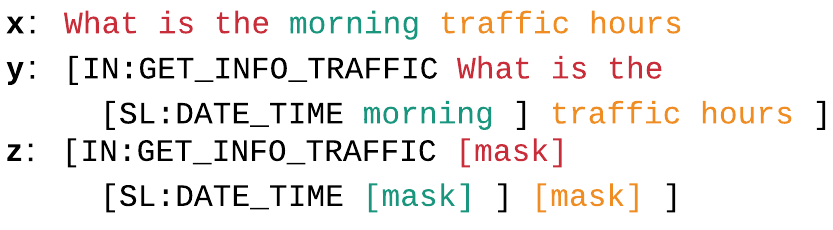}
    \caption{An example of an input utterance $\vx$, its desired output $\vy$, and the template $\vz$ inferred from $\vy$ . By definition, the template $\vz$ above can be used to generate other utterances such as ``\emph{how is the 5:00 traffic looking}'' or ``\emph{Any construction on my morning route}''.}
    \label{fig:xyz}
  \end{minipage}
  \hfill
  \begin{minipage}[b]{0.47\textwidth}
    \includegraphics[width=\textwidth]{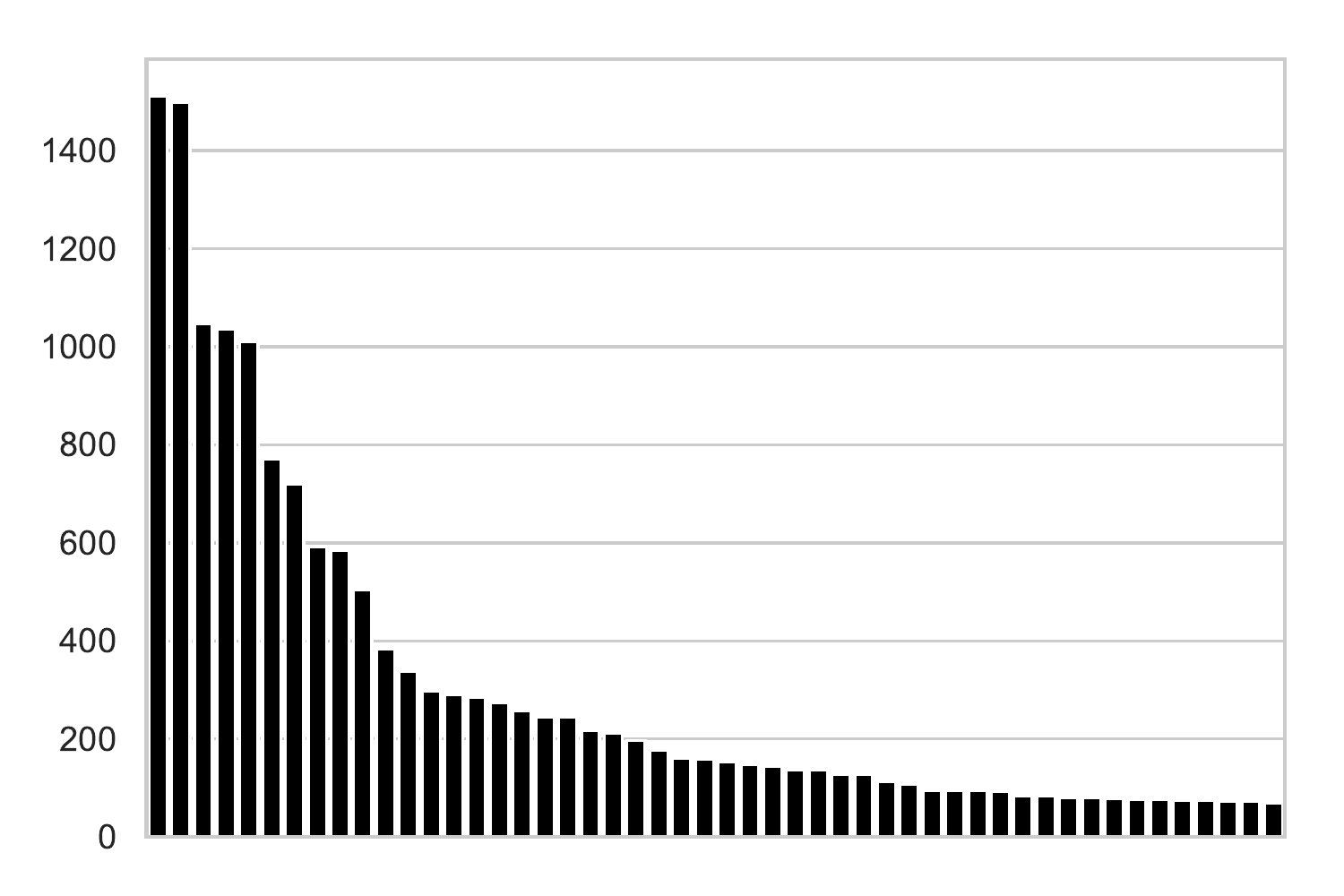}
    \caption{Frequency of  most 50 common templates in Facebook TOP dataset. The frequency of $\vz$ follows a power-law probability distribution.}
    \label{fig:treedist}
  \end{minipage}
\end{figure}

In this work, we investigate semantic parsing with hierarchical representations \citep{gupta-etal-2018-semantic-parsing} instead of the traditional logical forms \citep{Zettlemoyer2005}. Given an utterance $\vx$, our goal is to produce a tree-structured representation $\vy$ of the utterance where additional information about intents and slots is introduced at the non-terminal nodes of the tree. We define a \emph{template} $\vz$ of a given annotation $\vy$ as a result of replacing all terminal nodes by a generic \texttt{[mask]} node. Figure~\ref{fig:xyz}  shows an example of such an utterance $\vx$, its annotation $\vy$  and the corresponding template $\vz$.

The hierarchical representation for task-oriented parsing proposed in \citep{gupta-etal-2018-semantic-parsing} aims for ease of annotation and expressiveness. The dataset in their work, Facebook TOP, is the largest publicly available dataset in English for hierarchical semantic parsing. It has more than 44K annotated queries.
We look at the distribution of the templates in Facebook TOP and found that the dataset is highly unbalanced (Figure~\ref{fig:treedist}). The 10 most frequent templates account for 30\% of the training data and 14\% of the data are singletons, which are utterances with only a single occurrence.
This analysis suggests that it is beneficial to generate more synthetic data for templates with low frequencies. In the field of Natural Language Processing, using synthetic data via back-translation \cite{sennrich-etal-2016-improving} has shown a great success for machine translation \citep{edunov-etal-2018-understanding}. Unlike machine translation, generating synthetic data for hierarchical semantic parsing is less straightforward. Our work positions itself as one of the first to explore the possibility of generate text from graph (template) for semantic parsing.

In this paper, we propose a generic framework for augmenting a semantic parser with synthetic data. Our framework consists of two steps.
First, we train a generator, followed by top-$p$ sampling to generate diverse synthetic utterances conditioning on the above-mentioned templates. Generated utterances share similar hierarchical structures (\ie templates) with real training utterances while providing a wide spectrum of lexical variety. Second, we use an auxiliary parser for filtering on the generated candidates. The filtering step guarantees the quality of the synthetic data.
Our generator is a sequence to sequence (seq2seq) model that is pretrained on massive amount of monolingual data with text infilling objective (\S\ref{sec:bart}). We utilize BART \citep{lewis-etal-2020-bart}, a recently proposed denoising autoencoder, as our generator to avoid training it from scratch. The auxiliary parser can be arbitrary. We experiment with BART-based parser as well as state-of-the-art pointer network parser \citep[s2s-pointer;][]{rongali2020}.

The paper is structured as follows.  We introduce our generative model for synthetic data in Section~\S\ref{sec:bart}. Experimental results on Facebook TOP dataset and sub-sampled datasets to simulate low-resource scenario are presented in Section~\S\ref{sec:expr}. Section~\S\ref{sec:conclusion} concludes the paper.

\section{Denoising Sequence-to-Sequence as Generator}
\label{sec:bart}
The generative story for generating synthetic data $\mathcal{Y}_{\text{syn}}=\{ \widetilde{\vy}_i\}_{i=1}^M$
is given by
\begin{enumerate}
    \item draw a template $\vz\sim p_\phi(\vz)$;\footnote{During inference for generating synthetic data, we draw $\vz$ uniformly in order to generate more annotations for templates in the long tail.}
    \item draw an annotation $\vy\sim p_\theta(\vy\,|\,\vz)$ by filling each \texttt{[mask]} token in $\vz$ by a word or sequence of words from vocabulary $\mathcal{V}$;
\end{enumerate}
Note that the transformation from annotation $\vy$ to utterance $\vx$ is deterministic by removing non-terminals from $\vy$.
While $p_\phi(\vz)$ can be modeled by an autoregressive neural language model or a  Probabilistic Context Free Grammar \citep{johnson98}, in this work we sample template $\vz$ from seen templates in the data. We leave the possibility of generating new templates to future work.

\begin{figure}[h]
\begin{subfigure}{.48\textwidth}
  \centering
  \includegraphics[width=.8\linewidth]{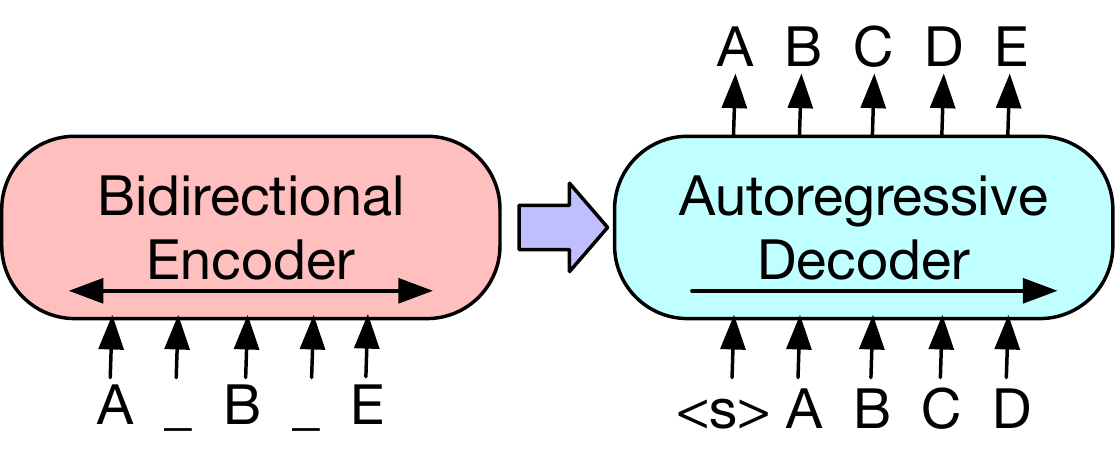}
  \caption{BART is trained to reconstruct the corrupted input.}
  \label{fig:bart}
\end{subfigure}
\par\bigskip
\begin{subfigure}{.48\textwidth}
  \centering
  \includegraphics[width=.8\linewidth]{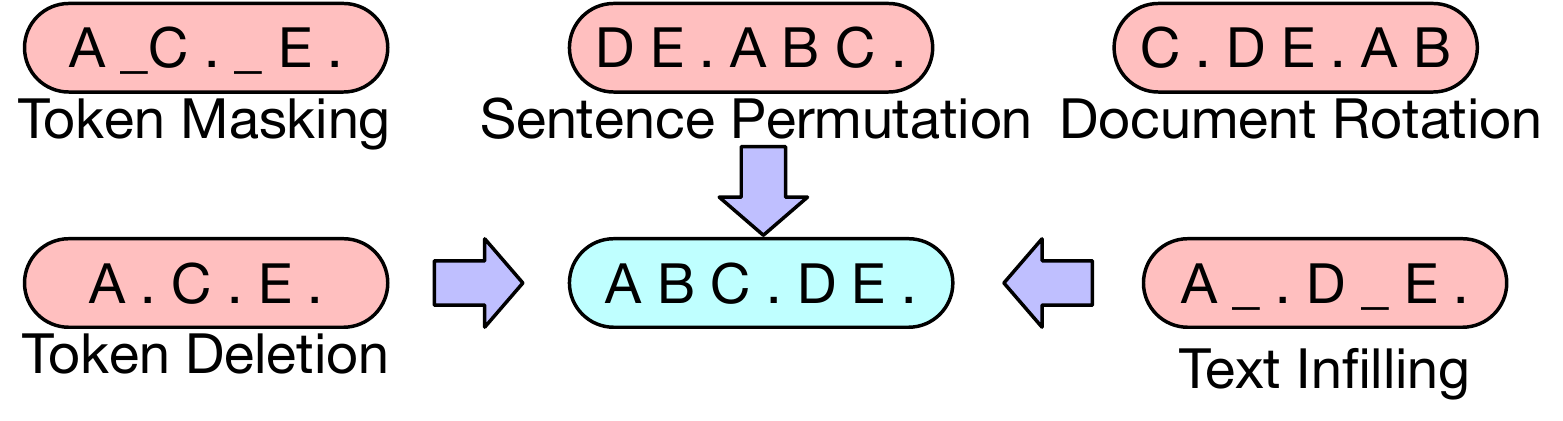}
  \caption{Five different types of noise introduced in BART.}
  \label{fig:noises}
\end{subfigure}
\caption{Overview of BART.}
\label{fig:bartoverview}
\end{figure}

We need a powerful conditional model $p_\theta(\vy\,|\,\vz)$ to generate annotation $\vy$. Thus, we choose BART, a pretrained denoising autoencoder for  sequence-to-sequence, as our model.
Figure~\ref{fig:bart} illustrates the idea behind BART. Given an input sequence (a stream of text), one of five types of noise  (Figure~\ref{fig:noises}) is used to corrupt the input sequence. Then BART reconstructs the original sequence by maximizing the likelihood of the original sequence.

\begin{figure*}
    \centering
    \includegraphics[width=\textwidth]{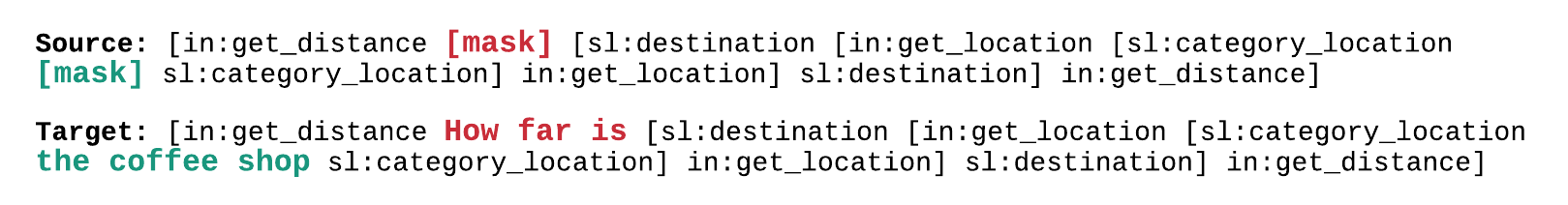}
    \caption{Data for fine-tuning BART.}
    \label{fig:format}
\end{figure*}

Since pretrained BART uses \emph{text infilling} as noise to corrupt the input sequence, naturally we can use BART to infill the templates. Text infilling is the task where a number of spans in the original input sequence are replaced by a token \texttt{[mask]} and BART is trained to predict the replaced spans in the position of \texttt{[mask]} tokens. For our purpose of generating synthetic data, we fine-tune BART on an infilling dataset where the input is a template $\vz$ with \texttt{[mask]}  and the output is a linearized tree representation $\vy$ where \texttt{[mask]} tokens are replaced by lexical words as shown in Figure~\ref{fig:format}.

{\bf BART source/target construction:} We call out a few processing steps to construct this infilling dataset. First, non-terminal words are lowercased. We find this is necessary since the input will be tokenized by BART tokenizer and lowercasing non-terminal words prevents over-segmentation. Second, we make each of the closing brackets ``\texttt{]}'' in the original data explicit (\eg \texttt{in:get\_distance]}, \texttt{sl:destination]}). This transformation provides the model explicit information of the scope of the intents and slots. 

{\bf Fine-tuning and generation:} We fine-tune BART generator using the (\textit{template}, \textit{annotation}) pairs. After fine-tuning, we use the generator to generate full parse trees given templates. To increase the diversity of generated samples, we use top-$p$ sampling \citep{Holtzman2020The} instead of beam search. The generator is trained to generate the tokenized labels together with the words. We remove generated annotations with invalid labels and convert the tokenized labels into the original tags in a post-processing step.

{\bf Auxiliary parser (AP) for filtering:} In our preliminary experiments, we found that the generated samples are noisy. When we train our parser on the concatenation of both real and generated samples, the test accuracy degrades by 1.13\% compared with a parser trained purely on real data.
We therefore use an auxiliary parser (AP) to select robust samples. The filtering step is straightforward. First, we train an auxiliary semantic parser $f_\theta(\vx)$ on the original Facebook TOP dataset. We then use this trained AP to parse synthetic data $(\widetilde{\vx}_i, \widetilde{\vy}_i)$ and keep those samples where the outputs of the parser $f_\theta(\widetilde{\vx}_i)$ match the synthetic labels $\widetilde{\vy}_i$ (i.e., $f_\theta(\widetilde{\vx}_i) = \widetilde{\vy}_i$).
The AP for filtering can be different from the target parser we train for semantic parsing. Therefore, we propose three settings: (1) BART as AP and a sequence-to-sequence model with pointer networks \citep[s2s-pointer;][]{rongali2020} as the target parser. (2) BART models for both AP and target parser. (3) s2s-pointer models for both AP and target parser. The comparisons and analysis are detailed in Section \S\ref{sec:expr}.

\section{Experiments}
\label{sec:expr}
We use Facebook TOP dataset in our experiments.
Statistics of the dataset are shown in Table~\ref{tab:top}. While there are more than 31K annotated utterances in training data, the number of unique templates is about 6K. As we have shown in Section~\ref{sec:intro}, the distribution of the templates is highly unbalanced. In training data, there are 1,511 annotations with the template  \texttt{[IN:UNSUPPORTED\_NAVIGATION [mask] ]} and 1,046 annotations with template \texttt{[IN:UNSUPPORTED [mask] ]}. In the case of no UNSUPPORTED setting ($-$ \texttt{UNSUPPORTED} in Table \ref{tab:results}), we exclude those annotations with \texttt{UNSUPPORTED} templates from train, valid, and test data. This results in 28,414 (template, annotation) pairs for training and 4,032 pairs for validation.

\begin{table}[h]
    \centering
    \begin{tabular}{@{} l c c c c @{}}
    \toprule
    Condition &&  \texttt{train} &  \texttt{valid} &  \texttt{test} \\
    \midrule
    $+$ \texttt{UNSUPPORTED} && 31,279 & 4,462 & 9,042 \\
    $-$ \texttt{UNSUPPORTED} && 28,414 & 4,032 & 8,241 \\
    \bottomrule
    \end{tabular}
    \caption{Number of samples in Facebook TOP dataset with ($+$) and without ($-$) \texttt{UNSUPPORTED} utterances.}
    \label{tab:top}
\end{table}
We fine-tune our BART generator using Adam optimizer \citep{adam} with a linear warmup of 4,000 steps at the peak learning rate of $2\mathrm{e}{-5}$. We pick the best model based on validation perplexity. After fine-tuning, we use the generator to sample 5 full parse trees per template.

\begin{figure*}
    \centering
    \includegraphics[width=\textwidth]{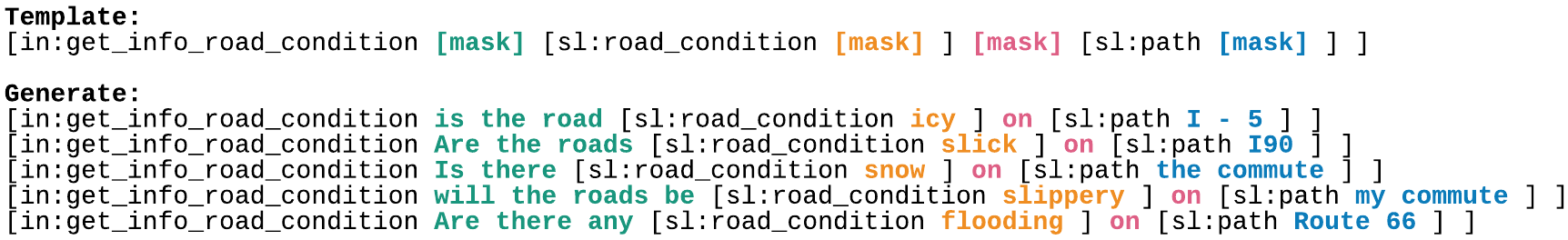}
    \caption{Sample of five synthetic parse trees generated given a template. Colors indicate the corresponding generated spans per \texttt{[mask]} token. The data is reformatted for readability.}
    \label{fig:synsamples}
\end{figure*}

\begin{table*}
    \centering
    \begin{tabular}{@{} l l l c l l c l @{}}
    \toprule
     &&&  \multicolumn{2}{c}{$-$ \texttt{UNSUPPORTED}}  && \multicolumn{2}{c}{$+$ \texttt{UNSUPPORTED}} \\
     \cmidrule{4-5} \cmidrule{7-8}
      Data &AP filter & Target parser & \#Samples & Acc (\%) && \#Samples & Acc (\%) \\
    \midrule
    \texttt{Real} && BART &  28,414 & 83.37 && 31,279 & 81.01 \\
    + \texttt{syn} & BART &  BART & 53,679 & 84.26 \textcolor{mygray}{(+0.89)} && 56,547 & 81.74 \textcolor{mygray}{(+0.73)} \\
    \midrule
    \texttt{Real} &&s2s-pointer& 28,414 & 84.80 && 31,279 & 82.10 \\
    \texttt{+syn} & BART & s2s-pointer & 53,679 & 85.31 \textcolor{mygray}{(+0.51)} && 56,355 & 82.71 \textcolor{mygray}{(+0.61)}\\
    \texttt{+syn} & s2s-pointer & s2s-pointer &  89,629 & 85.68 \textcolor{mygray}{(+0.88)} && 92,264 & 82.77 \textcolor{mygray}{(+0.67)} \\
    \bottomrule
  \end{tabular}
    \caption{Exact-match results of our experiments. The AP filter can be a fine-tuned BART for parsing or a s2s-pointer model of \citet{rongali2020}}
    \label{tab:results}
\end{table*}

The exact-match results for the three settings of using BART/s2s-pointer as auxiliary and target parser are given in Table~\ref{tab:results}. We first notice that the BART-based parser performs on-par with SOTA model based on pointer network and RoBERTa \citep{liu2020roberta} feature extractor in the work of \citet{rongali2020}. This suggests that pretraining a general purpose seq2seq model is beneficial for downstream conditional generation task. We also see that using synthetic data brings additional 0.89\% for BART-parser and 0.88\% for s2s-pointer parser on the exact-match accuracy. The gain of using synthetic data is smaller when \texttt{UNSUPPORTED} utterances are present in training and testing data.

Table~\ref{tab:acc_freq} shows the exact match accuracy of BART-based parser on testset with respect to template frequency $f$ in training data. We see that synthetic data helps low-frequency templates ($f<5$) the most (+1.36\%). The gain of 0.67\% for unseen templates ($f=0$) suggests that there is a room for further improvement by generating new templates.

\begin{table}[h]
  \centering
  \begin{tabular}{@{} l l r r r@{}}
  \toprule
  Training data &&  $f\ge 5$ &  $f<5$ & $f=0$\\
  \midrule
  \texttt{Real} && 89.46 & 74.70 & 61.90\\
  \texttt{+syn} && 90.30 & 76.06 & 62.57\\
   \midrule
  $\Delta$ && 0.84 & 1.36 & 0.67\\
  \bottomrule
  \end{tabular}
  \caption{Exact-match accuracy on testset with respect to template frequency $f$ in training data.}
  \label{tab:acc_freq}
\end{table}

In order to support new domains (with new intents and slots) for the virtual assistants, we investigate the role of synthetic data when there is a little data available for the new domains.
We simulate this scenario by sub-sampling 6K utterances in the training data as follows: for each template in the training data, we randomly choose one utterance. We use this sub-sampled data for training our parser, generator, and AP. Table~\ref{tab:lowresource} shows the mean and variance of the accuracy on five random sub-sampled portions of the train data. We see that in this low resource setting, our approach boosts the accuracy by more than 2\% absolute.

\begin{table}[ht!]
  \centering
  \begin{tabular}{@{} l c c @{}}
  \toprule
  Training data & \#Samples & Acc (\%)\\
  \midrule
  \texttt{Real} & \phantom{0}6,000 & 72.24 $\pm$ 0.05\\
  \texttt{+syn} &~30,000& \bf{74.31} $\pm$ 0.05 \\
  \bottomrule
  \end{tabular}
  \caption{Average accuracy of five different runs for 6K training examples. The synthetic data is filtered by BART parser, which is trained on 6K samples.}
  \label{tab:lowresource}
\end{table}

\section{Related Work}
Using pretrained models to generate synthetic data has been studied recently \citep{amin-nejad-etal-2020-exploring,kumar2020data}. Their work however focuses on multi-class classification problems. Taking a step further, our work shows a viable path for  structured output (\ie parse trees) problems.

\section{Conclusions}
\label{sec:conclusion}
We have proposed a novel approach for generating synthetic data for hierarchical semantic parsing. Our initial experiments show promising results of this approach and open up possibility for applying it to other problems with highly structured outputs in Natural Language Processing.

\section*{Acknowledgments}
We thank reviewers for their constructive comments and suggestions. We also thank Raquel G. Alhama for proofreading this paper.
\bibliography{amlc}
\bibliographystyle{acl_natbib}

\end{document}